\documentclass[letterpaper, 10 pt, conference]{ieeeconf}
\IEEEoverridecommandlockouts
\makeatletter
\let\NAT@parse\undefined
\makeatother

\usepackage{flushend}
\usepackage{hyperref}
\hypersetup{colorlinks=true, allcolors=black, urlcolor=[HTML]{0000EE}, linkcolor=red}
\usepackage{csquotes}
\usepackage{graphicx}
\usepackage{xcolor}
\usepackage{soul}
\usepackage{siunitx}
\usepackage[utf8]{inputenc}
\usepackage[compress]{cite}
\usepackage[T1]{fontenc}

\DeclareSIUnit\dBA{dBA}

\title{\LARGE \bf Acoustic Sensing for Universal Jamming Grippers}
\author{
Lion Weber$^{*,1,2}$ \and Theodor Wienert$^{*,1}$ \and Martin Splettstößer$^{*,1,2}$ \and Alexander Koenig$^{1,3}$ \and Oliver Brock$^{1,2,3}$ \\
    \thanks{$^{*}$ Equal contributions} 
    \thanks{$^{1}$ Robotics and Biology Laboratory, Technische Universität Berlin}
    \thanks{$^{2}$ Science of Intelligence, Research Cluster of Excellence, Berlin} 
    \thanks{$^{3}$ Robotics Institute Germany}
    \thanks{
 This paper extends our workshop contribution~\cite{weber2025acoustic} by evaluating sensor resolution across object orientations and providing an extensive discussion of the approach. Portions of the earlier version, including text and figures, are reused or adapted in this manuscript.
 }
    \thanks{
 This work has been partially supported by the German Federal Ministry of Research, Technology and Space (BMFTR) under the Robotics Institute Germany (RIG) and funded by the Deutsche Forschungsgemeinschaft (DFG, German Research Foundation) under Germany's Excellence Strategy -- EXC 2002/1 -- project no. 390523135; and project no. 405033880.
 }
    \thanks{Project page with videos and code: \url{https://rbo.gitlab-pages.tu-berlin.de/papers/acoustic-jamming-icra26}}
}

\begin{document}

\maketitle
\begin{abstract}
Universal jamming grippers excel at grasping unknown objects due to their compliant bodies. Traditional tactile sensors can compromise this compliance, reducing grasping performance.
We present acoustic sensing as a form of \textit{morphological sensing}, where the gripper's soft body itself becomes the sensor. A speaker and microphone are placed inside the gripper cavity, away from the deformable membrane, fully preserving compliance. Sound propagates through the gripper and object, encoding object properties, which are then reconstructed via machine learning.
Our sensor achieves high spatial resolution in sensing object size (\SI{2.6}{\milli\meter} error) and orientation (\SI{0.6}{\degree} error), remains robust to external noise levels of \SI{80}{\dBA}, and discriminates object materials (up to \SI{100}{\percent} accuracy) and $16$ everyday objects (\SI{85.6}{\percent} accuracy).
We validate the sensor in a realistic tactile object sorting task, achieving $53$ minutes of uninterrupted grasping and sensing, confirming the preserved grasping performance. Finally, we demonstrate that disentangled acoustic representations can be learned, improving robustness to irrelevant acoustic variations.

\end{abstract}

\section{Introduction}

Universal jamming grippers~\cite{brown_universal_2010} grasp by enveloping an object with a flexible membrane filled with granular media, and then stiffening through jamming. 
This simple, compliant mechanism lets them conform to a wide variety of unknown objects, making them particularly attractive for grasping in unstructured environments. Despite their versatility, universal jamming grippers remain difficult to sensorize: traditional rigid sensors compromise compliance, reducing grasping performance. Yet, tactile sensing is essential in realistic environments. Acoustic sensing offers a solution that preserves compliance while providing rich tactile feedback.

Acoustic sensing is a form of \textit{morphological sensing}~\cite{wall24Thesis}, where the gripper's soft body itself functions as the sensor. The key idea is that interactions between the gripper and environment induce measurable changes in the gripper's morphology. Sound propagating through the gripper captures these changes as modulated signals, from which machine learning reconstructs desired sensor measurements.

\begin{figure}
    \centering
    \includegraphics[width=\linewidth]{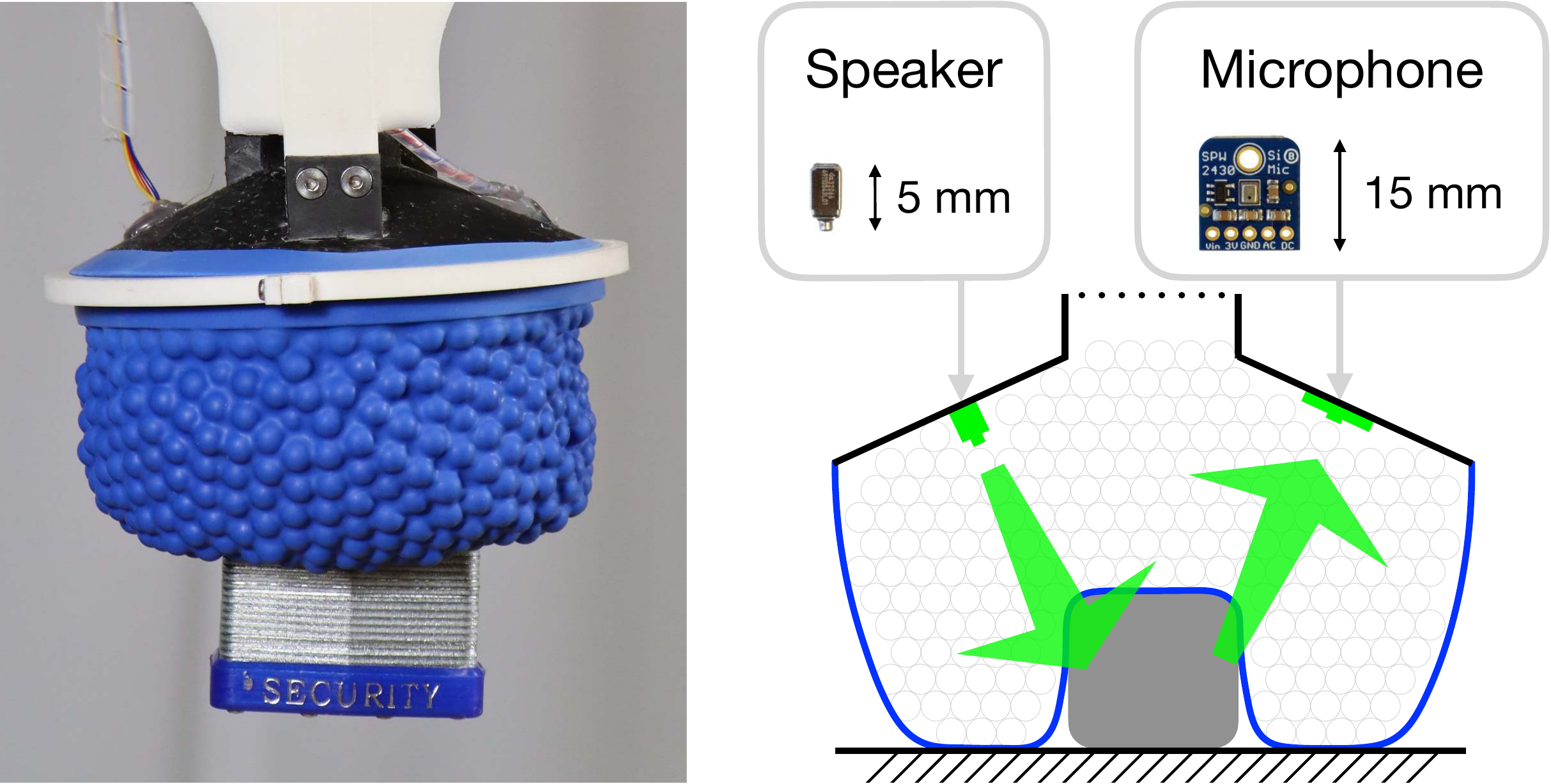}
    \caption{
 Left: Universal jamming grippers conform to unknown objects, enabling versatile grasping, but traditional sensors restrict this deformability. Right: We address this with acoustic sensing. A speaker emits sound into the gripper cavity, which propagates through the gripper and object, encoding object properties (size, material, orientation, class, etc.) in the modulated signal shown in green. A microphone then records the signal, and machine learning reconstructs the object state. 
 }
    \label{fig:overview}
\end{figure}

In this paper, we explain how to build an acoustic sensor that fully preserves the gripper's compliance (Fig.~\ref{fig:overview}). We then show that a single acoustic sensor can perceive many physical quantities that may otherwise require specialized sensors. 
The acoustic sensor estimates object \textit{size} with millimeter-scale resolution and remains robust to substantial external acoustic noise. It reliably senses object \textit{materials}, which is impossible for cameras when objects appear visually identical. Our approach also senses the object \textit{orientation} with sub-degree resolution.
In a realistic application, we discriminate the \textit{class} of $16$ household objects with a high accuracy of \SI{85.6}{\percent} and show the gripper's unaffected grasping performance in a tactile object-sorting task. Finally, we demonstrate that the high-dimensional acoustic data can be transformed into disentangled latent spaces, isolating task-relevant information and improving robustness against irrelevant variations such as object pose.

\section{Related Work}

We first analyze existing sensing methods for universal jamming grippers, with a focus on their impact on gripper compliance, and then present prior art on acoustic sensing. 

\subsection{Sensorized Universal Jamming Grippers}

Some prior works on sensorizing universal jamming grippers rely on mounting rigid cameras on or inside the gripper. For example, a small camera can be placed at the center of the gripper membrane~\cite{meng_universal_2023}, but the rigid mount inside the gripper likely worsens compliance. Other efforts relying on cameras~\cite{li_tata_2022, sakuma_universal_2018, sakuma_jamming_2023} fill the gripper cavity with liquid and glass beads of the same refractive index. Cameras mounted inside the gripper can then see the membrane deformation through the glass beads to estimate the object's shape and pose. A challenge of these approaches is protecting the electronics inside the gripper and other robot parts from liquid leakage. Most of these works~\cite{sakuma_universal_2018, sakuma_jamming_2023} require custom membranes with physical markers, which were shown to worsen grasping capabilities~\cite{sakuma_universal_2018} and are complex to manufacture compared to using latex balloons as membranes~\cite{brown_universal_2010}. 

Other works directly sensorize the gripper membrane with flexible sensor arrays~\cite{loh_3d_2021,mo_empowering_2024}. These sensors can predict proximity, membrane deformation, normal forces~\cite{loh_3d_2021}, and slippage~\cite{mo_empowering_2024}. However, the attached sensor arrays again limit the membrane's compliance, require a complex manufacturing process, and have coarse spatial resolution of 6$\times$6~\cite{loh_3d_2021} or 3$\times$3~\cite{mo_empowering_2024} sensor grids. Our sensor is mounted far away from the membrane and therefore preserves its compliance, is easy to manufacture, and can accurately reconstruct various physical quantities from a single sensor signal.

\subsection{Acoustic Sensing}

Industry and robotics research increasingly leverage acoustic signals. In industrial settings, acoustic sensing detects tool wear~\cite{liang1989toolwear}, rail infrastructure failure~\cite{lee2016Rails}, and motor defects~\cite{GLOWACZ2019106226}. In robotic manipulation with rigid grippers, acoustic sensing classifies household objects~\cite{sinapov_interactive_2009, xu2025}, estimates object materials and contact positions~\cite{lu_active_2023, xu2025}, and enables peg-in-hole insertion~\cite{zhang2025vibecheck}. Acoustic sensing is also applied in soft robotic grasping. Wall et al.~\cite{wall_IJRR_2021} present an acoustic sensor integrated into soft robotic fingers, which accurately predicts contact locations, forces, materials, and temperature. They demonstrated that \textit{active} acoustic sensing, where sound is injected rather than passively observed, yields more robust predictions~\cite{wall_IJRR_2021}. Other works reconstruct the deformed shape of soft robotic actuators using acoustics~\cite{Yoo2024, Sofla2024}. While prior work demonstrated the versatility of acoustic sensing across many robotics domains, it has not yet been applied to universal jamming grippers---despite these grippers being challenging to sensorize due to their deformable nature. To our knowledge, we are the first to bridge this gap.

\section{Acoustic Sensing for Universal Jamming Grippers}

This section explains how to build and operate the acoustic jamming gripper, showing how the gripper's deformable body, which is normally a challenge for sensorization, turns into an \textit{asset} for morphological sensing.

\subsection{The Gripper Body as a Sensor}

Our design goal is to augment the gripper with sensing capabilities while fully preserving its compliance. Acoustic sensing offers a solution: the hardware, a speaker and microphone, can be placed away from the deformable membrane, leaving the compliant morphology untouched~(Fig.~\ref{fig:overview}). But acoustic sensing does more than just preserve compliance---it exploits it. The same deformability that lets the gripper conform to objects for a stable grasp also shapes the acoustic signal in a way that reveals object properties. Acoustic sensing leverages the same morphology as a resonant structure: morphology enables \textit{grasping}---and it enables \textit{sensing} too. When the soft jamming gripper conforms to an object, the resulting imprint in the cavity encodes information about the object's geometry and pose relative to the gripper. The large contact area between gripper and object facilitates energy transfer \textit{into} the object, revealing intrinsic object properties such as material composition. The microphone captures the reflected sound waves, modulated by both the altered cavity and the object itself. Subsequent computation then extracts object properties embedded in this modulation.

\subsection{Building an Acoustic Jamming Gripper}

Our implementation builds on the classic universal jamming gripper design~\cite{brown_universal_2010}. 
A latex balloon filled with granular medium is mounted on a rigid conical housing and connected to a vacuum source~(Fig.~\ref{fig:overview}). At ambient pressure, the balloon deforms around the object, conforming to its shape.  Applying negative pressure jams the granular medium, stiffening the structure and securing the grasp. The gripper design files, all data, and software are available in our code repository.

\begin{figure}
    \centering
    \includegraphics[width=0.8\linewidth]{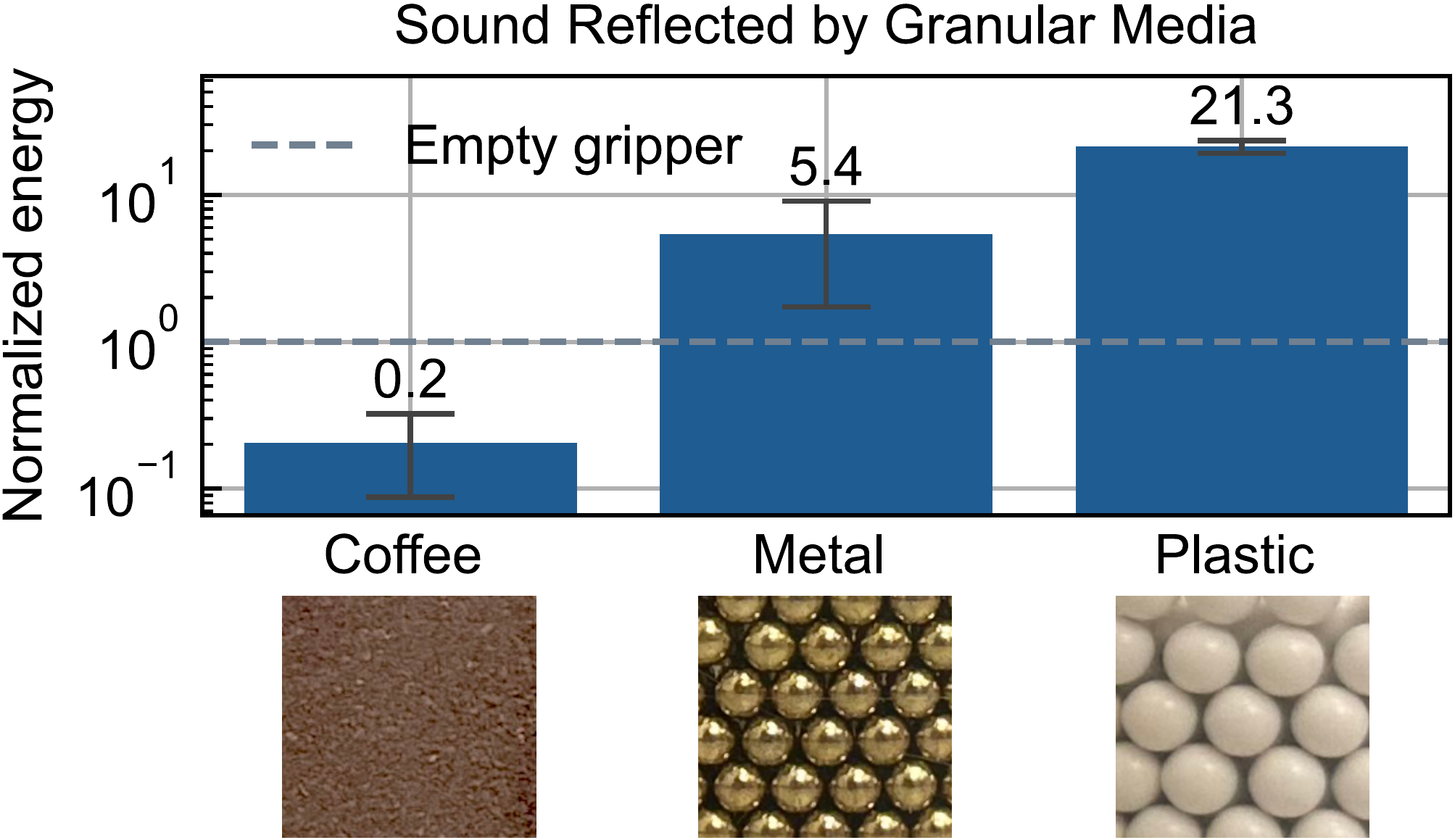}
    \caption{Different granular media inside the gripper have different sound absorption characteristics. Good sound reflection is necessary to return information about the environment to the microphone. The plot shows the energy of the sound signal as reflected by different granular media inside the gripper. Results are scaled relative to the sound reflection in an empty gripper (i.e., when only air is inside). We use plastic ball bearings as our granular medium because they better reflect the speaker's sound than ground coffee and metal ball bearings.}
    \label{fig:energy}
\end{figure}

Installing the electronics on the jamming gripper is simple. We mount a small speaker (Knowles RAB-32063-000) and microphone (Adafruit SPW2430) inside the cavity. Wires are routed through \SI{2}{\milli\meter} holes in the housing and sealed with hot glue. The speaker and microphone are positioned \SI{7}{\centi\meter} apart and angled \SI{25}{\degree} toward the gripper's contact region to maximize acoustic transmission and reception while avoiding direct coupling~(Fig.~\ref{fig:overview}, right). Importantly, both elements are rigidly fixed to the housing rather than the membrane, ensuring that the deformable body remains unimpeded. The microphone signal is amplified using a small preamplifier (Adafruit PAM8302A) and digitized through an 18-bit analog-to-digital converter at \SI{44.1}{\kilo\hertz} (MAYA44 USB+). The speaker signal is generated via the 20-bit digital-to-analog converter of the same interface. All supporting electronics, including amplification and audio interface, are located outside the gripper. The total cost of the sensing electronics is below $140$~USD. Overall, this demonstrates the setup is low-cost, simple to integrate, and fully compatible with the soft, deformable structure of the gripper.

Acoustic sensing relies on the reflection of sound: the gripper filling must transmit and reflect energy so the microphone can capture signals modulated by the object. Therefore, selecting the right granular medium is crucial as it determines how effectively sound can travel towards the object and back to the microphone. Fig.~\ref{fig:energy} shows that coffee grounds, commonly used in jamming grippers~\cite{brown_universal_2010}, attenuate more than \SI{80}{\percent} of the acoustic energy and are therefore unsuitable for sensing. This result is in line with previous research showing that coffee has strong sound-insulating properties~\cite{kang_investigation_2023}. We compare metallic and plastic ball bearings as alternative fillings. Ball bearings of plastic (\SI{6}{\milli\meter} diameter, \SI{0.2}{\gram}, Elite Force WA41839) and metal (\SI{4.5}{\milli\meter} diameter, \SI{0.36}{\gram}, Walther Premium Steel BBs 4.1668-1) reflect the sound better. We chose plastic ball bearings because they provide the best sound reflection, while metal ball bearings are heavy and can short-circuit electronics.

\subsection{Using an Acoustic Jamming Gripper}

\begin{figure}
    \centering
    \includegraphics[width=\linewidth]{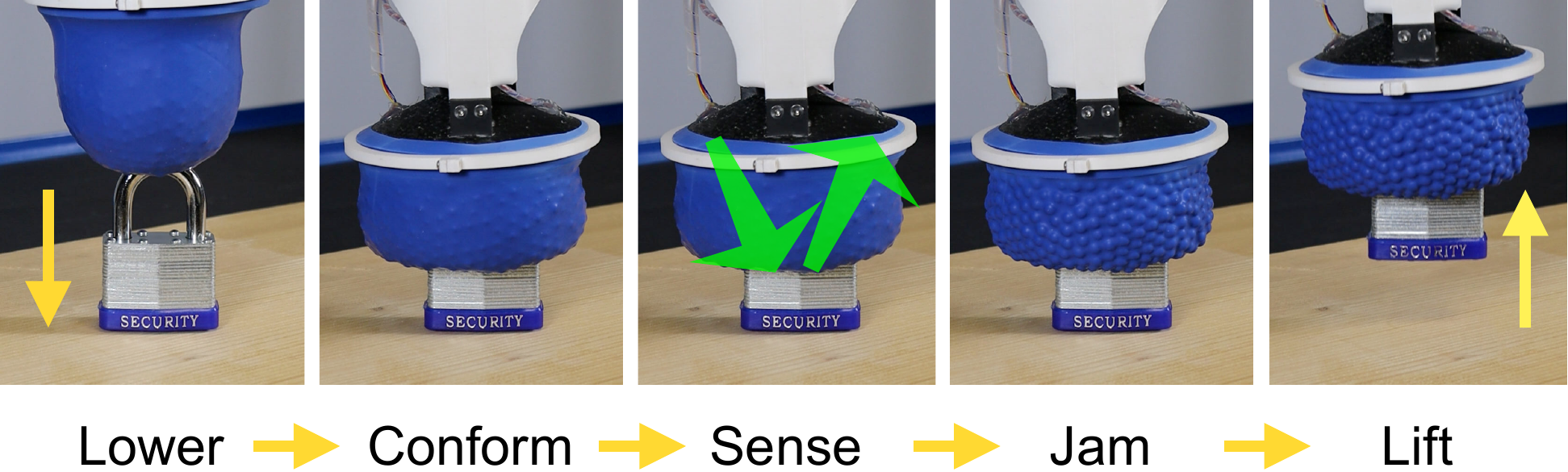}
    \caption{ 
 The sensing and grasping process. We sense after conforming to the object, when the contact area for acoustic energy exchange is large, but before jamming, because the lower air pressure during evacuation reduces sound transmission in the cavity and the vacuum pump generates substantial noise. The supplementary video shows the grasping process.
 }
    \label{fig:grasping_process}
    \vspace{-0.1cm}
\end{figure}

Fig.~\ref{fig:grasping_process} shows the proposed sensing and grasping process. First, we loosen the medium and approach the object with an impedance controller on a Franka Panda arm. The gripper conforms to the object until an ATI Mini40 sensor on the gripper mount reaches a contact force of \SI{8}{\newton}. We then keep the gripper in position. We allow two seconds for the settling of the granular medium and for venting air that is displaced during the compression of the cavity. The remaining contact force facilitates vibration exchange between the granular medium and the object. The sensing process happens at ambient pressure and not while jamming because the negative pressure reduces sound transmission through air, and the vacuum pump induces vibrations that propagate into the cavity. We play a one-second logarithmic frequency sweep from \SI{20}{\hertz} to \SI{20}{\kilo\hertz} on the speaker to emphasize lower frequencies, which tend to travel farther than high frequencies. The microphone records for one second as the sound plays. After reconstructing the object properties from the reflected audio signal, we jam the medium by evacuating the gripper, and lift the object.

We process the one-second microphone recordings following the approach by Wall et al.~\cite{wall_IJRR_2021}. We compute a short-time Fourier transform (STFT) with a window size of $2048$ to obtain $1025$ frequency bins. We sum the STFT output across the time axis and normalize to get our final $1025$-dimensional feature vector. This representation encodes the morphological imprint of the object, including cues about object properties. To decode these morphological cues, we employ a three-layer convolutional neural network (CNN) followed by a fully connected layer.
This last layer maps to a regression value or a vector of class probabilities via a softmax operation, depending on the task. Our loss is a mean-squared error for regression and cross-entropy for classification. We use an $80$:$20$ train-validation split in a $5$-fold cross-validation. All results in this paper report the validation mean and variance from this cross-validation unless indicated otherwise.

\begin{figure}
    \centering
    \includegraphics[width=\linewidth]{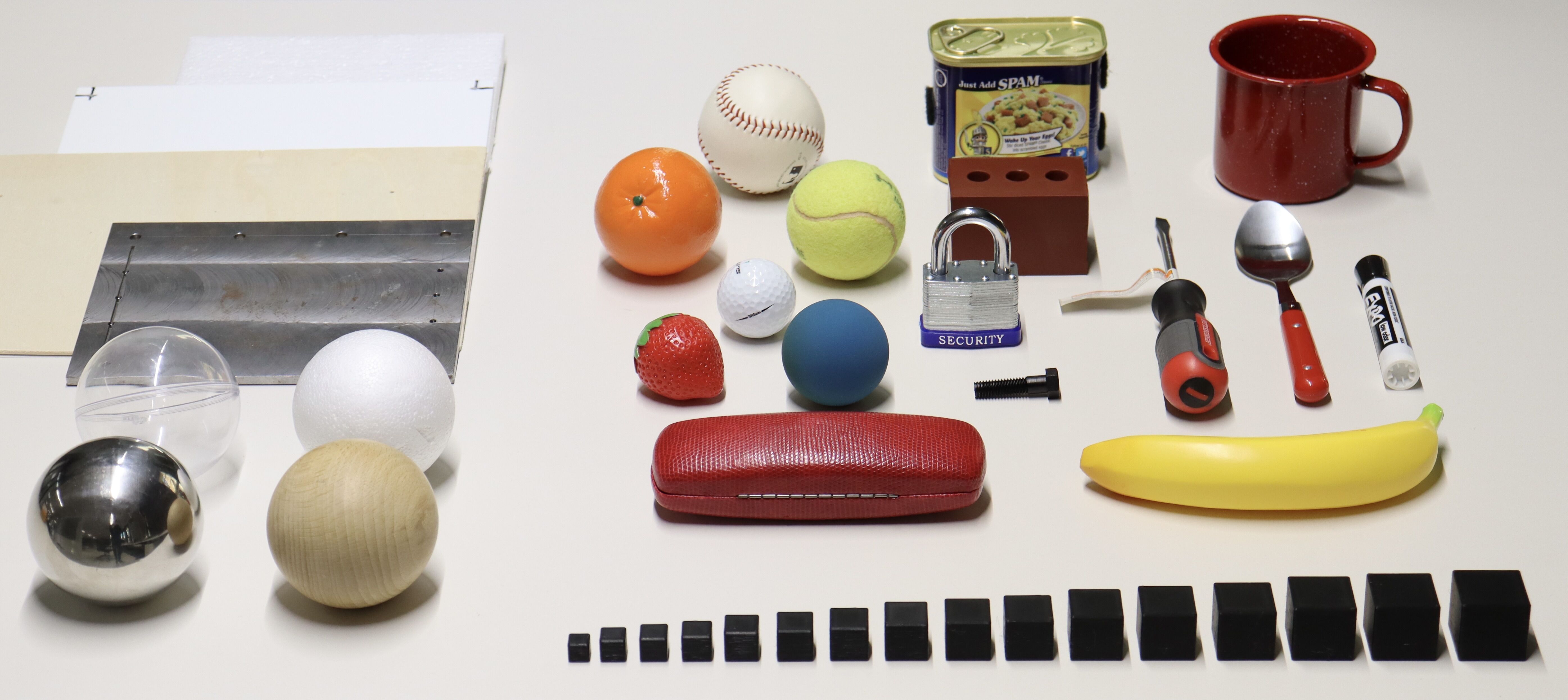}
    \caption{The object set used in this paper (\textit{bottom}: cubes of different sizes for Section~\ref{sec:sizes}; \textit{left}: plates and spheres of different materials for Section~\ref{sec:material}; \textit{right}: everyday objects from the YCB~\cite{ycb} dataset for Section~\ref{sec:obj_class})}
    \label{fig:objects}
\end{figure}
\begin{figure}
    \centering
    \includegraphics[width=0.8\linewidth]{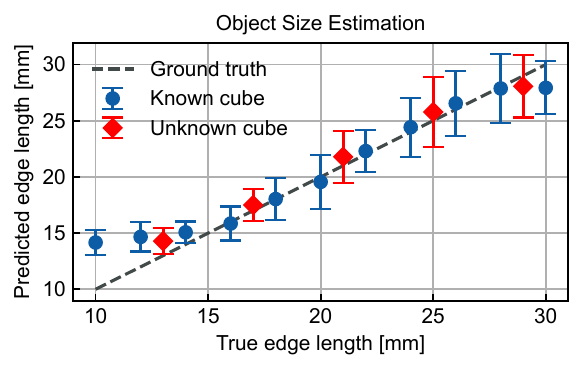}
    \caption{The acoustic sensor has high spatial resolution: it can accurately predict the size of enclosed objects. Our model predicts the cube edge length with a small root mean square error (RMSE) of \SI{2.7}{\milli\meter} for known objects, \SI{2.4}{\milli\meter} for unknown objects, and an overall error of \SI{2.6}{\milli\meter}. Fig.~\ref{fig:objects} shows the corresponding cubes. }
    \label{fig:size}
\end{figure}
\section{Sensor Evaluation \label{sec:sensor_eval}}

This section evaluates the sensor's resolution, robustness to external acoustic noise, and discriminative performance across quantities, including object size, material, and orientation---tasks that typically require specialized sensors.

\subsection{Evaluating Sensor Resolution Across Object Size\label{sec:sizes}}

In this experiment, we evaluate the spatial resolution of the acoustic sensor by distinguishing objects of slightly different sizes. We 3D-print $16$ cubes with edge lengths ranging from \SI{10}{\milli\meter} to \SI{30}{\milli\meter} using Polylactic Acid~(PLA), keeping material density and shape constant (Fig.~\ref{fig:objects}). Fig.~\ref{fig:size} shows predictions for both known and unknown cubes, with the latter not seen during training. The sensor achieves a small overall error of \SI{2.6}{\milli\meter}. The largest error occurs for the smallest cube, likely because the acoustic modulations are weak as the cube size (\SI{10}{\milli\meter}) approaches the granular medium diameter (\SI{6}{\milli\meter}).

How does this sensor resolution compare to related work? Camera-based tactile sensing for jamming grippers can achieve high, pixel-level spatial resolution~\cite{li_tata_2022}. However, approaches relying on flexible sensor arrays attached to the gripper membrane presumably have lower spatial accuracy, on the order of a few centimeters. For example, Mo et al.~\cite{mo_empowering_2024} use a 3$\times$3 tactile array on an \SI{80}{\milli\meter}$\times$\SI{80}{\milli\meter} membrane. They do not explicitly report spatial resolution, but this layout implies an effective minimum resolution of roughly \SI{27}{\milli\meter}. By this measure, our acoustic sensor provides an order-of-magnitude higher spatial resolution while fully preserving gripper compliance and maintaining a simple design.
\begin{figure}
    \centering
    \includegraphics[width=0.71\linewidth]{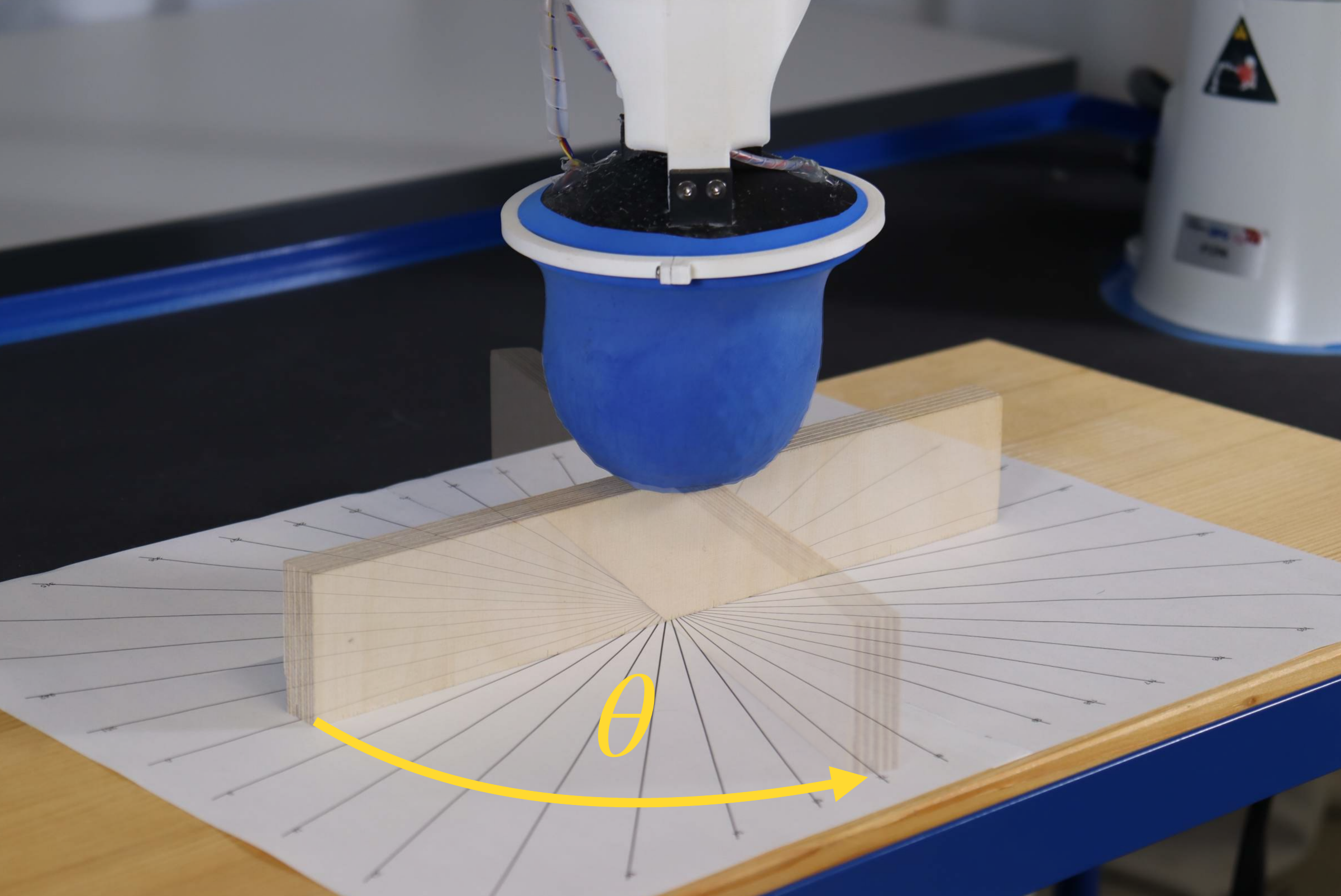}
    \caption{We predict the orientation of a wooden bar from acoustic data alone. The angle $\theta$ parameterizes the orientation. We sample data every \SI{9}{\degree} across \SI{180}{\degree}, summing to $19$ distinct object rotations. The size of the bar is \SI{270}{\milli\meter}$\times$\SI{15}{\milli\meter}$\times$\SI{44}{\milli\meter}. The figure overlays two photos of different object orientations. Fig.~\ref{fig:pose_prediction_results} shows the results of the experiment.}
    \label{fig:pose_prediction_exp_setup}
\end{figure}

\begin{figure}
    \centering
    \includegraphics[width=0.8\linewidth]{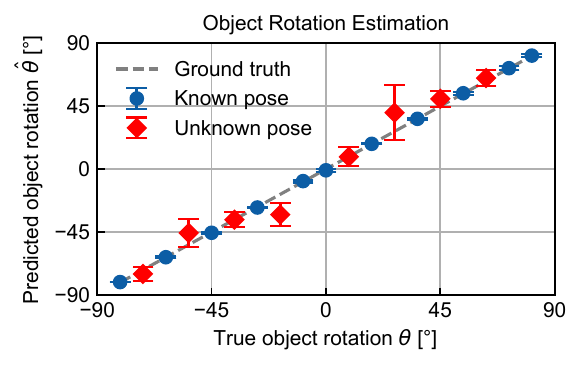}
    \caption{The acoustic sensor accurately senses an object's orientation. We predict the rotation of a wooden bar inside the gripper as parametrized by an angle $\theta$. 
 The average validation error across known object orientations is \SI{0.6}{\degree} and the test error on orientations not seen at training is \SI{8.0}{\degree}.
 }
    \label{fig:pose_prediction_results}
\end{figure}

\subsection{Evaluating Sensor Robustness to External Acoustic Noise}

Universal jamming grippers are used in unstructured environments and may be exposed to external acoustic noise. Therefore, we test the generalization of the trained object size predictor to external noise. We place a consumer-grade loudspeaker at approximately \SI{40}{\centi\meter} from the gripper, play white noise, and measure the noise level at the gripper using a mobile phone. We then reevaluate the sensor's capability to measure object size. Our results show that the RMSE rises from \SI{2.7}{\milli\meter} at \SI{45}{\dBA} ambient noise levels to \SI{3.5}{\milli\meter} at \SI{80}{\dBA} external noise. The small error increase of \SI{0.8}{\milli\meter} under substantial noise is surprising but aligns well with the results by Wall et al.~\cite{wall_IJRR_2021}, which also showed good robustness against external noise. The membrane seems to act as a good insulator for external noise, making the introduced noise inside the cavity relatively smaller than the reflected speaker signal level. 

\begin{figure}
    \centering
    \includegraphics[width=0.8\linewidth]{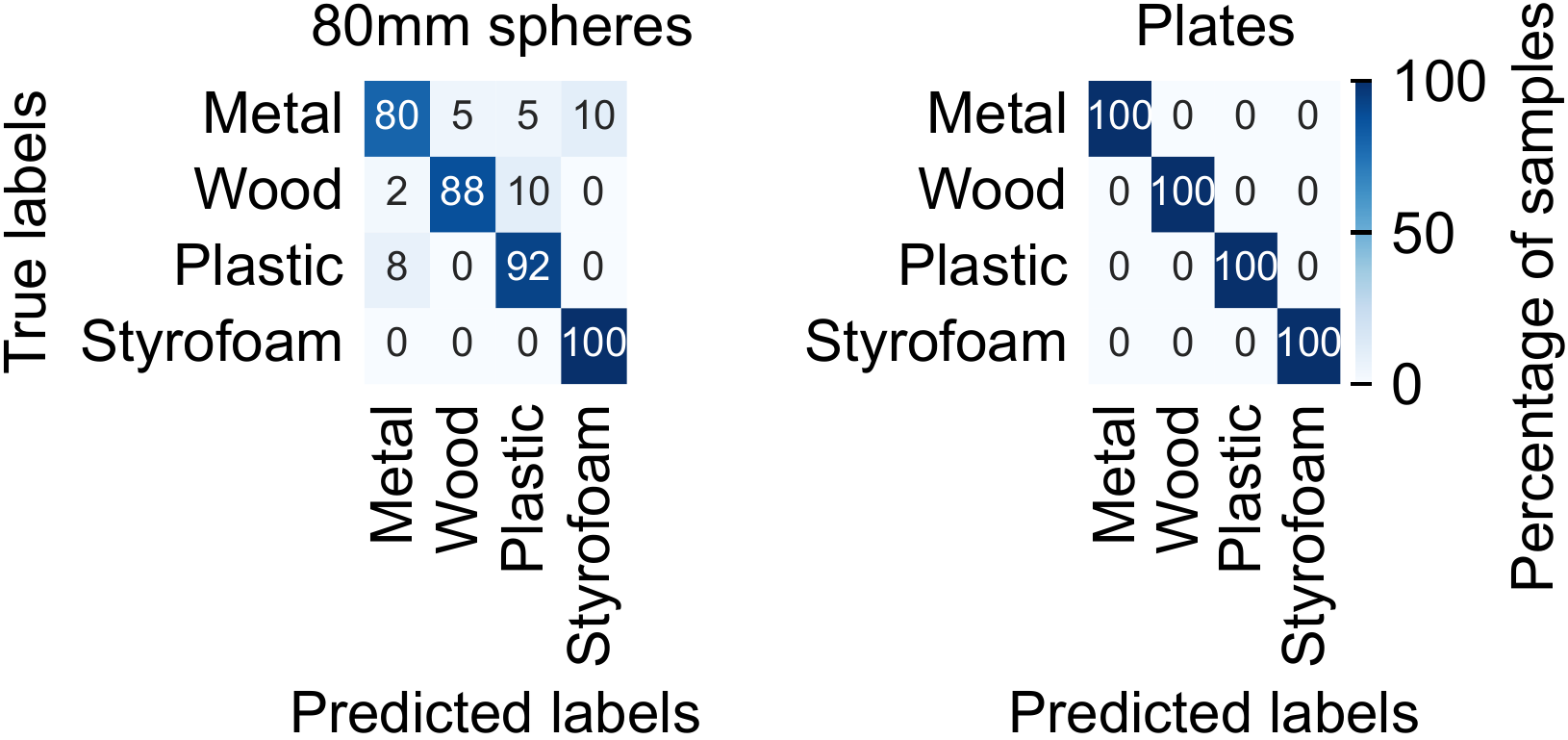}
    \caption{The acoustic sensor detects the materials of \SI{80}{\milli\meter} spheres with a high accuracy of \SI{90}{\percent} (left). It perfectly classifies plates of different materials (right). Fig.~\ref{fig:objects} shows the corresponding objects. We place these objects on a concave rubber piece, preventing the spheres from rolling.}
    \label{fig:materials_plates}
\end{figure}

\subsection{Evaluating Sensor Resolution Across Object Orientation\label{sec:pose}}  

Sensing the object orientation within the gripper immediately before grasping is challenging for vision-based methods due to self-occlusion. Therefore, we evaluate the acoustic sensor's resolution across another dimension: object orientation. In the experiment in Fig.~\ref{fig:pose_prediction_exp_setup}, we record acoustic data of an object in $19$ unique rotations and attempt to predict its orientation as represented by an angle $\theta$. As shown in Fig.~\ref{fig:pose_prediction_results}, our method achieves high-resolution orientation estimation with a validation error of only \SI{0.6}{\degree}. Notably, we withhold $8$ of the $19$ object rotations for testing---an unusually large test split of \SI{42}{\percent}---yet the model still generalizes well to unseen rotations with a test error of \SI{8.0}{\degree}. These results demonstrate that the acoustic sensor can achieve high-resolution orientation estimation, complementing its size and material sensing capabilities.

\subsection{Evaluating Sensor Discrimination of Object Materials\label{sec:material}} 

We hypothesized that part of the speaker's acoustic energy propagates \textit{into} the object and reflects back into the gripper cavity toward the microphone. Therefore, the sensor should be able to discriminate \textit{intrinsic} object properties such as material. To test this, we classify \SI{80}{\milli\meter} spheres and $\sim$\SI{5}{\milli\meter} thick plates made of metal, wood, plastic, and Styrofoam (Fig.~\ref{fig:objects}). The results in Fig.~\ref{fig:materials_plates} show that the spheres are distinguished with \SI{90}{\percent} accuracy, while plates can be perfectly differentiated with \SI{100}{\percent} accuracy.

An additional experiment with smaller, \SI{30}{\milli\meter} spheres yielded accuracies near random chance. Because material information lies \textit{inside} the object, energy must be transferred into the object and back to the gripper. Smaller spheres have a reduced contact area with the membrane, limiting this energy transfer, possibly explaining the lower classification performance. In contrast, object size prediction on small objects (\SI{10}{\milli\meter} to \SI{30}{\milli\meter}) worked well, suggesting that predicting size relies less on energy exchange between object and gripper. Presumably, object size can be estimated directly from the object's imprint in the granular medium.

In summary, acoustic sensing allows accurate material prediction for sufficiently large objects. This capability is particularly useful in unstructured environments, because cameras cannot differentiate visually identical objects that differ in other physical properties like material.

\begin{figure}
    \centering
    \includegraphics[width=0.95\linewidth]{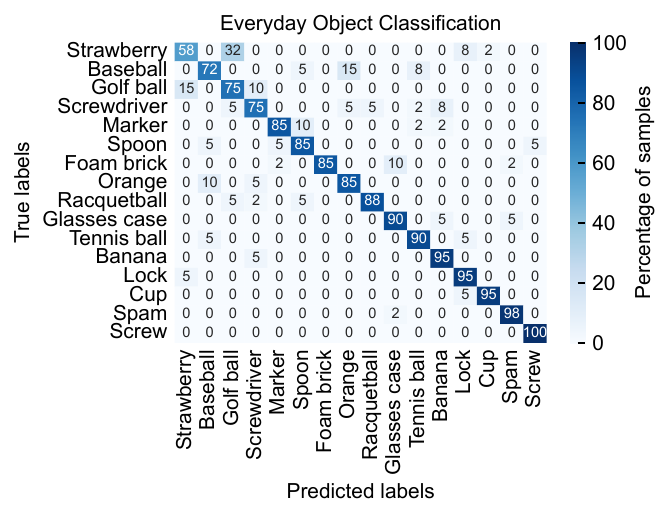}
    \caption{
 After showing strong performance on simple shapes, the sensor also performs well on realistic objects: it classifies $16$ everyday objects from the YCB~\cite{ycb} dataset with an avg. validation accuracy of \SI{85.6}{\percent}. Some objects with similar shape and size are occasionally confused (e.g., strawberry and golf ball, or baseball and orange). Fig.~\ref{fig:objects} shows the corresponding objects.}
    \label{fig:ycb}
\end{figure}

\section{Sensor Application in Realistic Setting \label{sec:obj_class}}

The previous section showed the efficacy of the acoustic jamming gripper in predicting various physical quantities of simple objects like cubes, plates, and spheres. This section applies and discusses the approach in a realistic setting when classifying complex, everyday objects.

\subsection{Application to Object Classification and Sorting\label{sec:object_classification}}

In this section, we demonstrate the system's applicability in a real-world sensing and grasping task. A natural choice is a purely tactile object sorting task, commonly found in industrial settings where vision is impaired due to occlusion. First, we need a more diverse object dataset including objects of different size, shape, and softness. The YCB~\cite{ycb} dataset is suitable because it provides a wide range of objects along exactly these dimensions. However, not all objects in this dataset fit inside the gripper. Therefore, we select some objects that fit inside the gripper and are also graspable. Fig.~\ref{fig:objects} shows the $16$ everyday household objects we selected from the larger YCB dataset. Our object set features deformable objects (foam brick, tennis ball, racquetball), small objects (screw), large objects (spam can, glasses case), and objects of similar shape (golf ball, racquetball, baseball, tennis ball, plastic orange, plastic strawberry). 

\begin{figure}
    \centering
    \includegraphics[width=0.85\linewidth]{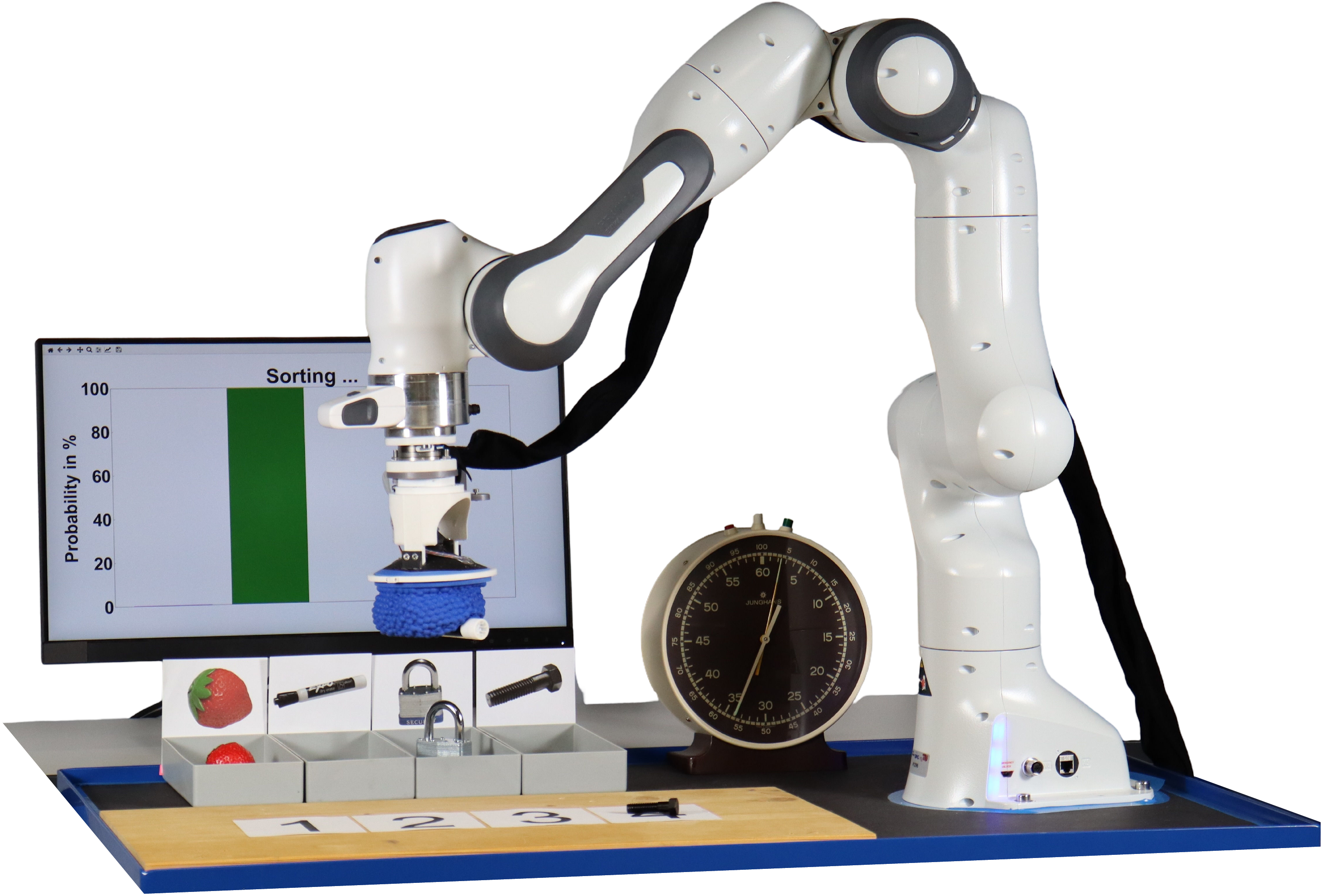}
    \caption{Our purely tactile object sorting task shows that while objects are classified correctly from sound, the sensing hardware fully preserves the gripper's compliant grasping. The supplementary video shows the robot reliably sensing the objects' class, grasping, and sorting them into the correct bin. The dashboard on the screen visualizes the object probabilities.}
    \label{fig:sorting}
    \vspace{-0.05cm}
\end{figure}

In realistic environments, object pose uncertainty may be large. To make our models robust against object pose changes, we record data in 20 random gripper poses for each object ($\pm$~\SI{90}{\degree} $z$-rotation and $\pm$ \SI{1}{\centi\meter} translation). Since audio signal variance between poses is large but samples within the same pose are nearly identical, we record only two one-second samples per pose to avoid overfitting. A dataset for $16$ objects consists of $\sim$~11 minutes of audio ($16\textrm{ objects}\times 20\textrm{ poses}\times 2\textrm{ samples}\times\SI{1}{\second}=\SI{640}{\second}$), taking roughly one hour to record. 
Fig.~\ref{fig:ycb} shows that our object classifier achieves \SI{90}{\percent} average validation accuracy or more for seven objects and \SI{85.6}{\percent} overall. The sensor only confuses some objects of similar geometry and size (e.g., strawberry and golf ball, or baseball and orange).

Finally, we demonstrate the system's strong sensing and grasping performance in a real-world tactile object sorting task. Fig.~\ref{fig:sorting} shows a still from the supplementary video. The video highlights that the acoustic sensor fully preserves the gripper's compliance and grasping capabilities while objects are classified correctly purely based on sound. We achieve $53$ minutes of uninterrupted sorting of four objects: strawberry, marker, lock, and screw. The operator places the objects in a randomized order at known grasping locations. The gripper then approaches, classifies, and sorts each object into its corresponding bin. We complete $39$ successful sorts in a row, each requiring a correct prediction, a stable grasp, and a controlled release. The demonstration ends at the 40th classification because no object probability crosses the user-defined minimum confidence threshold of \SI{60}{\percent} as lock and strawberry receive similar class probabilities. Nevertheless, this last prediction would have also been correct. The video also demonstrates robustness to slight variations in pose as the operator sets down the objects in poses that are likely not observed during training. The gripper always manages to secure a stable grasp and ensures no object drops. In summary, our results imply that the acoustic sensor can support differentiating objects in realistic, cluttered environments such as warehouses, where reliance on vision is not always possible due to occlusions or insufficient lighting.

\subsection{Data Requirements and Model Baselines}

The practical deployment of any learning-based robotic sensing system is constrained by the cost of data collection. To better understand the trade-offs in our acoustic sensing approach, we analyze data requirements and compare model architectures using the largest and most diverse dataset recorded in this work: the $16$-object YCB set.

\begin{figure}[t]
    \centering
    \includegraphics[width=0.8\linewidth]{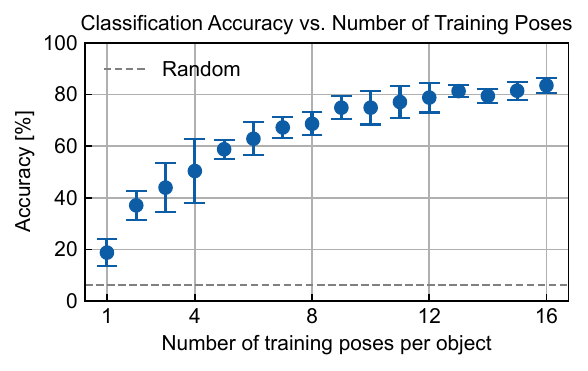}
    \caption{Reaching adequate YCB~\cite{ycb} object classification accuracies of roughly \SI{80}{\percent} requires sampling training data in at least 12 different poses per object. Our best model trains on data from 16 different poses.}
    \label{fig:acc_over_train_set_size}
\end{figure}

We first investigate how much training data is required to achieve good classification performance. Our dataset contains $20$ poses per object ($16$ training, $4$ validation poses), with two samples per pose. In Fig.~\ref{fig:acc_over_train_set_size}, we evaluate how classification accuracy changes as we vary the number of unique poses per object in the training set. The results show that we need 12 gripper poses to reach adequate classification accuracies of approximately \SI{80}{\percent}. While the performance improves sharply when increasing from very few poses, adding more poses continues to yield gains but with diminishing returns. This analysis provides practical guidance for the effort required to collect training data.

\begin{figure}
    \centering
    \includegraphics[width=0.8\linewidth]{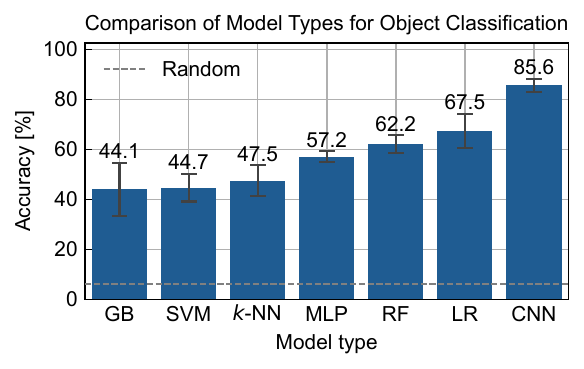}
    \caption{
 Classifier performance on raw STFT input (\textit{GB}: gradient boosting, \textit{SVM}: support vector machine with RBF kernel, \textit{$k$-NN}: $k$-nearest neighbors, \textit{MLP}: multi-layer perceptron, \textit{RF}: random forests, \textit{LR}: multi-class logistic regression, \textit{CNN}: our best-performing convolutional neural network)}
    \label{fig:different_models}
\end{figure}

Next, we evaluate different classifiers on the same STFT feature vectors to understand the impact of model choice. Fig.~\ref{fig:different_models} shows that our CNN achieves the highest accuracy of \SI{85.6}{\percent}, averaged over all cross-validation folds, clearly outperforming other baselines. Presumably, the lower performance of the other classifiers is partly due to the high dimensionality of the input features ($1025$ dimensions). In such high-dimensional input spaces, distance metrics lose their meaning due to the curse of dimensionality. Therefore, methods based on computing distances (such as $k$-NN and SVM~\cite{cortes1995support}) suffer from the reduced discriminative power of distance metrics. While tree-based and linear models (RF, GB, LR) perform better than random chance, the CNN still clearly outperforms them.

\subsection{Limitations} 

We now discuss the limitations of the acoustic sensing approach. 
Acoustic data is high-dimensional and complex, making analytical modeling of these phenomena challenging. We therefore employ machine learning to uncover patterns in the data. This is the strength of the approach, but also its weakness, as it inherits the limitations of machine learning. 

For instance, the approach is data-dependent as the models presented so far are supervised. Hence, new training data is required for new object classes or modalities we wish to predict, but collecting data in the real world is costly. A mitigation would be to learn more general object representations---such as object shape---to directly analyze novel objects without task-specific datasets. The success of vision and language models was mainly due to the availability of vast amounts of diverse training data. However, the presented acoustic sensing approach is highly morphology-specific, which may limit scalability. For example, collecting data cannot be easily parallelized on multiple grippers, because cross-gripper transfer remains challenging.

Like all machine learning models, our approach is susceptible to shortcut learning, where the model exploits spurious correlations in the training data rather than the true causal features. For instance, in our object orientation prediction experiments in Section~\ref{sec:pose}, we initially rotated the gripper to automate data collection instead of rotating the object. This caused the model to overfit to the sound of the arm's motors and fail to generalize to actual object rotations. By rotating the object during training data collection, we achieved good generalization, highlighting the importance of careful dataset collection to avoid unintended shortcuts.

Moreover, our models perform well on a given gripper; however, model transfer to \textit{different} gripper morphologies remains challenging. For instance, replacing or repositioning the latex balloon alters the acoustic properties of the cavity, which can reduce the reliability of trained models. The core difficulty is that \textit{any} morphological change introduces an acoustic distribution shift: when caused by object contact, such shifts carry useful tactile information, but when caused by gripper design changes, they instead introduce an undesired acoustic shift. Because the models cannot separate these cases, they do not transfer well across gripper designs. These results align with the work by Wall~et~al.~\cite{wall_IJRR_2021}, who report that models transfer poorly between different soft robot fingers. 

We do not view sensitivity to changes in gripper morphology as a fundamental challenge but as an \textit{entanglement} problem. The reflected sound jointly encodes multiple factors: object state (size, material, orientation), gripper morphology (membrane placement, filling, jamming state), and environmental conditions (ambient noise, contact force, pump vibrations). So far, we have shown that variations in \textit{one} of these factors---the object state---can be reconstructed via morphological sensing. In principle, with appropriate data and inductive biases, machine learning should be able to disentangle \textit{all} of these generative factors. This perspective motivates the search for disentangled acoustic representations, which we explore in the next section.

\subsection{Disentangled Representations for Morphological Sensing}

The acoustic signal reflects multiple entangled factors, including object state, gripper morphology, and environmental conditions. Disentangled representations aim to factorize these variations, aligning each axis in a latent space with a semantically meaningful factor. Invariant subspaces---such as pose-invariant or gripper-invariant subspaces---can be understood as slices through this higher-dimensional disentangled space, where irrelevant axes are marginalized. Learning such lower-dimensional representations is desirable because they isolate task-relevant variations (e.g., object properties) while ignoring irrelevant ones (e.g., object pose or gripper design), enabling robustness to such irrelevant factors.

We show how such representations can be discovered without semantic labels by exemplifying how a pose-invariant subspace can be found. Using self-supervised dimensionality reduction~\cite{Hadsell2006}, we map measurements of the same object in different poses close together in the latent space, while separating measurements of different objects. This approach requires no object class labels, highlighting a practical advantage: disentangled representations can be learned directly from the data without expensive labeling.

\begin{figure}
    \centering
    \includegraphics[width=\linewidth]{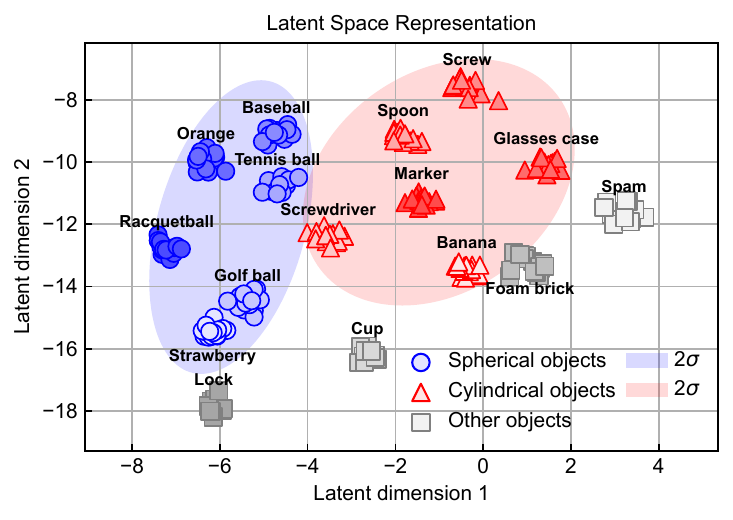}
    \caption{Disentangled representations can be learned from acoustic data without class labels. The learned latent space separates object properties from pose: objects with similar shape cluster together regardless of pose---e.g., \textit{spherical} objects (orange, baseball, tennis ball, racquetball, golf ball, strawberry) and \textit{cylindrical} objects (screw, spoon, glasses case, marker, screwdriver, banana). 
    The shaded ellipses denote Gaussian covariance contours at \SI{2}{\sigma}. Each data point corresponds to a sample in a different object pose. 
    This plot shows training data points for better readability; Figs.~\ref{fig:distance-matrix} and \ref{fig:knn-after-self-sup} provide quantitative results on validation data.}
    \label{fig:dim-red}
    \vspace{0.1cm}
\end{figure}

\begin{figure}
    \centering
    \includegraphics[width=0.8\linewidth]{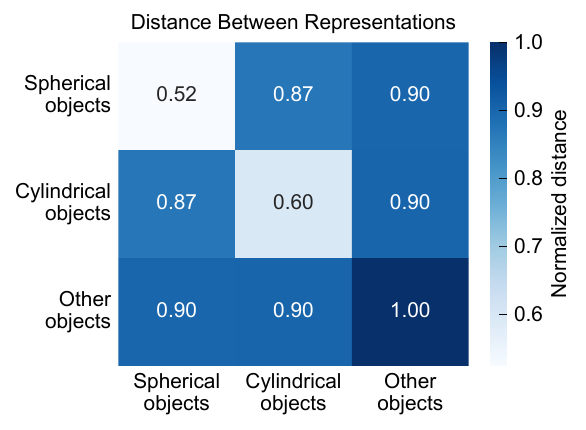}
    \caption{Our quantitative analysis confirms the semantic structure in the latent space: intra-category distances for \textit{spherical} and \textit{cylindrical} objects are smaller than inter-category distances, showing that the representations cluster objects by shape. \textit{Other} objects span diverse geometries and therefore remain well separated, as reflected in their large intra-category distance. While Fig.~\ref{fig:dim-red} is a snapshot of one fold, we average over all folds of the $k$-fold cross-validation here, providing a more comprehensive view across the entire dataset and different initializations.}
    \label{fig:distance-matrix}
\end{figure}

\begin{figure}
    \centering
    \includegraphics[width=0.8\linewidth]{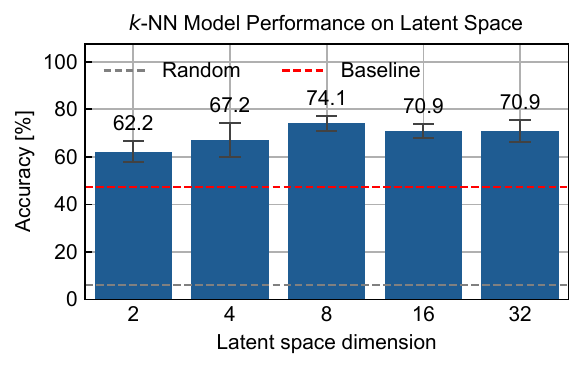}
    \caption{
 Disentangled embeddings improve object separability: applying $k$-NN on the latent space yields up to \SI{26.6}{\percent} higher accuracy than on raw STFT features, showing that the learned representations expose relevant variations while ignoring irrelevant factors like pose.
 The $k$-NN baseline is \SI{47.5}{\percent} from Fig.~\ref{fig:different_models}. Performance peaks at eight latent dimensions, likely reflecting a balance between capturing all meaningful variations in the acoustic signal while avoiding overfitting to irrelevant factors.}
    \label{fig:knn-after-self-sup}
\end{figure}

Our experiments show that the latent space in Fig.~\ref{fig:dim-red} captures pose-invariant object properties: each object occupies a distinct region regardless of pose and a semantic structure emerges. \textit{Spherical} and \textit{cylindrical} objects gather in common regions, and spherical objects of similar size cluster together (e.g., orange, baseball, and tennis ball group together, as do golf ball and strawberry), while \textit{other} objects with diverse properties cover different parts of the space. 
The analysis in Fig.~\ref{fig:distance-matrix} confirms this semantic structure: intra-category distances for \textit{spherical} and \textit{cylindrical} objects are smaller than inter-category distances. Objects with diverse geometries, such as those in the \textit{other} category, are correctly kept apart in the latent space, reflected in their large intra-category distance. 
Finally, Fig.~\ref{fig:knn-after-self-sup} shows the latent space improves separability of the acoustic data: a simple $k$-NN classifier on the disentangled embeddings achieves up to \SI{26.6}{\percent} higher accuracy than on raw STFT features, confirming that the latent space exposes meaningful semantic structure while ignoring task-irrelevant variations, such as object pose.

In principle, the same approach can be extended to discover other invariant subspaces. For example, a gripper-invariant subspace can be obtained by ensuring that variations in gripper morphology (e.g., membrane placement, filling, etc.) do not change the embedding for the same object. Achieving this requires collecting repeated measurements across different gripper configurations. Overall, these findings demonstrate that disentangled subspaces exist within the high-dimensional acoustic data and that they can be discovered from data alone. By appropriately constraining the learning process, one can extract subspaces that are invariant to specific nuisance factors, enabling more robust and transferable acoustic sensing.

\section{Conclusion}

We demonstrated that the soft body of a universal jamming gripper, typically a challenge for sensorization, can be turned into an asset through morphological sensing. In this framework, the gripper's morphology acts as a resonant body, capturing object properties as acoustic variations.

We made three key contributions. First, we showed that acoustic sensing is a suitable morphological sensing approach for jamming grippers, as the rigid hardware can be placed away from the deformable parts, fully preserving the gripper's compliance. Our realistic object sorting task confirmed that grasping performance is maintained, achieving 53 minutes of purely tactile sorting without dropping a single object.
Second, we demonstrated that a single sensor can accurately reconstruct various object properties, including size, material, orientation, and class, with predictions robust to external noise due to the shielding effect of the gripper membrane.
Third, we showed that disentangled representation spaces exist within the high-dimensional acoustic data. These latent spaces separate task-relevant information, such as object properties, from irrelevant variations, like object pose, enabling more robust predictions. Overall, our work calls for a perspective shift: a robot's body is not merely a passive structure, but an active participant in perception, realized through morphological sensing.


\bibliographystyle{IEEEtran}
\bibliography{references}

\end{document}